\pdfoutput=1

\documentclass[11pt]{article}

\usepackage{acl}

\usepackage{times}
\usepackage{latexsym}

\usepackage[T1]{fontenc}

\usepackage[utf8]{inputenc}

\usepackage{microtype}

\usepackage{amsmath}
\usepackage{array}
\usepackage{enumitem}
\usepackage{subcaption}
\usepackage{pifont}
\usepackage{multirow}
\usepackage{graphicx}

\usepackage{float}
\usepackage{xspace}
\usepackage{caption}
\usepackage{booktabs}
\newcommand{\dataset}[1]{\textit{#1}\xspace}
\newcommand{\teach}{\dataset{TEACh}}
\newcommand{\teachda}{\dataset{TEACh-DA}}

\newcommand{\commander}{\textit{Commander}}
\newcommand{\follower}{\textit{Follower}}
\newcommand{\robot}{\textit{Follower}}

\newcommand{\task}[1]{\textsc{#1}}

\newcolumntype{L}[1]{>{\raggedright\let\newline\\\arraybackslash\hspace{0pt}}m{#1}}
\newcolumntype{C}[1]{>{\centering\let\newline\\\arraybackslash\hspace{0pt}}m{#1}}
\newcolumntype{R}[1]{>{\raggedleft\let\newline\\\arraybackslash\hspace{0pt}}m{#1}}

\usepackage[colorinlistoftodos,prependcaption,textsize=tiny]{todonotes}

\usepackage{marginnote}

\title{Dialog Acts for Task-Driven Embodied Agents}

\author{ Spandana Gella\thanks{\hspace{0.5em}These two authors contributed equally.}, Aishwarya Padmakumar$^*$, Patrick Lange, Dilek Hakkani-Tur \\
Amazon Alexa AI \\
  \texttt{\{sgella,padmakua,patlange,hakkanit\}@amazon.com} \\}

\begin{document}
\maketitle

\begin{abstract}
Embodied agents need to be able to interact in natural language -- understanding task descriptions
and asking appropriate follow up questions to obtain necessary information to be effective at successfully accomplishing tasks for a wide range of users.
In this work, we propose a set of dialog acts for modelling such dialogs and
annotate the \teach dataset that includes over 3,000 situated, task oriented conversations (consisting of 39.5k utterances in total) with dialog acts. 
\teachda is one of the first large scale dataset of dialog act annotations for embodied task completion.
Furthermore, we demonstrate the use of this annotated dataset in training models for tagging the dialog acts of a given utterance, predicting the dialog act of the next response given a dialog history, and use the dialog acts to guide agent's non-dialog behaviour.
In particular, our experiments on the \teach Execution from Dialog History task where the model predicts the sequence of low level actions to be executed in the environment for embodied task completion, 
demonstrate that dialog acts can improve end task success rate by up to 2 points compared to the system without dialog acts.
\end{abstract}


    
    

\section{Introduction}

Natural language communication has the potential to significantly improve the accessibility of embodied agents. 
Ideally, a user should be able to converse with an embodied agent as if they were conversing 
with another person and the agent should be able to understand tasks specified at varying levels of 
abstraction and request for help as needed, identifying any additional information that needs to be obtained in follow up questions. 
Human-human dialogs 
that demonstrate such behavior are critical to the development of effective human-agent communication.  
Annotation of such dialogs with dialog acts is beneficial to better understand common conversational situations an agent will need to handle~\cite{gervits2021hurdl}.
Dialog acts can also be used in building task oriented dialog systems to plan how an agent should react to the current situation~\cite{dstcseries}. 

In this paper, we design a dialog act annotation schema for embodied task completion based on the dialogs of the \teach dialog corpus~\cite{padmakumar2021teach}. 
\teach is a dataset of over 3,000 situated text conversations between human annotators role playing a user (\commander) and a robot (\robot) collaborating to complete household tasks such as making coffee and preparing breakfast in a simulated environment. 
The tasks are hierarchical, resulting in agents needing to understand task instructions provided at varying levels of abstraction across dialogs. 
The human annotators had a completely unconstrained chat interface for communication, so the dialogs reflect natural conversational behavior between humans, not moderated by predefined dialog acts or turn taking. 
Additionally, the \follower\ had to execute actions in the environment that caused physical state changes which were examined to determine whether a task was successfully completed.
We believe that these annotations will enable the study of more realistic dialog behaviour in situated environments, unconstrained by turn taking.

Summarizing our contributions:
\begin{itemize}[noitemsep]
    \itemsep0em 
    \item We propose a new schema of dialog acts for task-driven embodied agents. This consists of 18 dialog acts capturing the most common communicative functions used in the \teach dataset. 
    \item We annotate the \teach dataset according to the proposed schema to create the \teachda dataset.
    \item We investigate the use of the proposed dialog acts in an extensive suite of tasks related to language understanding and action prediction for task-driven embodied agents.
\end{itemize}

We establish baseline models for classifying the dialog act of a given utterance in our dataset and predicting the next dialog act given an utterance and conversation history.
Additionally, we explore whether dialog acts can aid in plan prediction - predicting the sequence of object manipulations the agent needs to make to complete the task, and Execution from Dialog History (EDH) - where the agent predicts low level actions that are executed in the virtual environment and directly evaluated on whether required state changes were achieved.


\begin{figure*}[t]
    \centering
    \includegraphics[width=.99\textwidth]{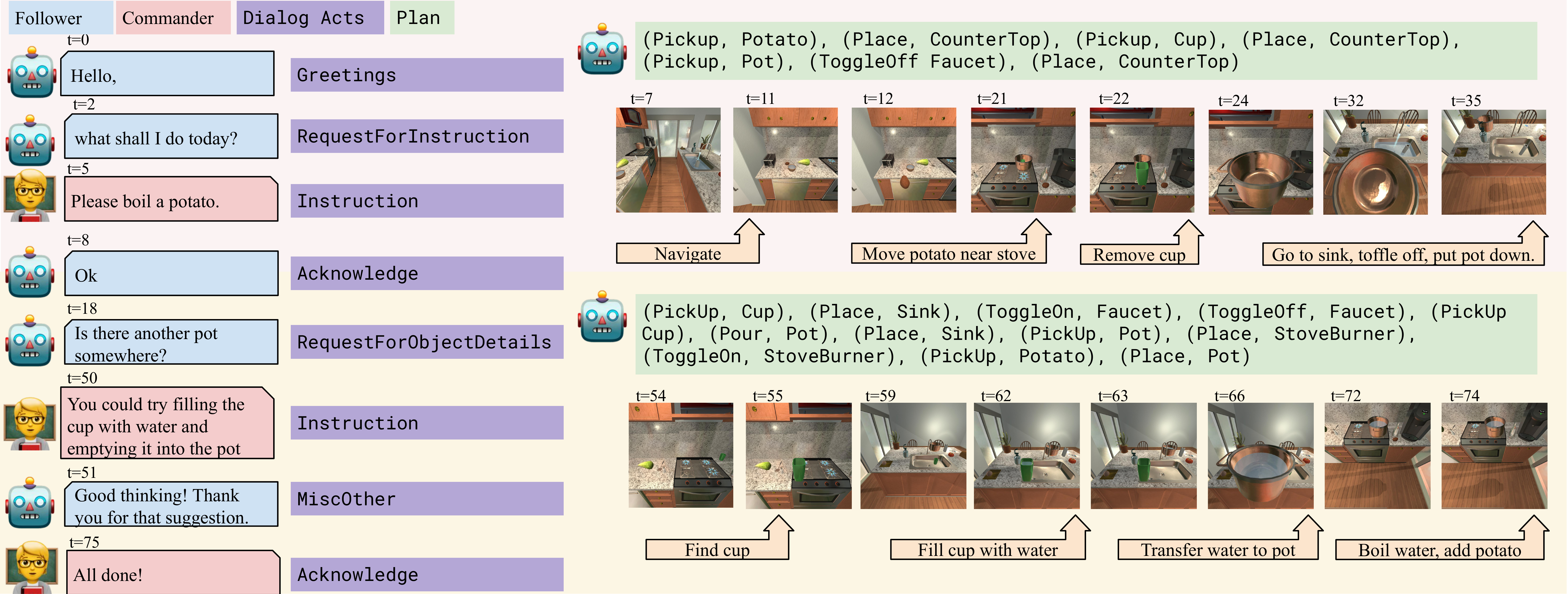}
    \caption{ Illustration of example session for the task \textit{Boil Potato} with corresponding 
    dialog acts for each utterance and plans with corresponding actions in the game session.} 
    \label{fig:example_figure_da}
\end{figure*}


\section{Related Work}

Dialog act annotations are common in language-only task-oriented dialog datasets, and are commonly used to plan the next agent action in dialog management or next user action in user simulation~\cite{dstcseries,multiwoz,fbtod,atis,doc2dial,taskmaster}.
Many frameworks have been proposed to perform such annotations. Some examples are DAMSL (Dialog Act Markup in Several Layers) and ISO (International Organization for Standardization) standard~\cite{Core1997CodingDW,syoung2007,Bunt2009TheDT,MezzaCSTR18}. 
Such standardization of dialog acts across applications has been shown to be beneficial for improving the performance of dialog act prediction models~\cite{MezzaCSTR18,PaulGH19}.

Most task-oriented dialog (TOD) applications and dialog act coding standards assume that the tasks to be performed can be fully specified in terms of slots whose values are  entities~\cite{syoung2007}.
However, we find that if we need to adopt a slot-value scheme for multimodal task-oriented dialog datasets such as \teach, much of the information that needs to be conveyed is not purely in the form of entities. For example, If an utterance providing a location of an object: 
``the cup is in the drawer to the left of the sink'' is to be coded at the dialog act level simply as an \texttt{INFORM} act, it could for example have a slot value called \texttt{OBJECT$\_$LOCATION} but the value of this would need to refer to most of the utterance, i.e. ``the drawer to the left of the sink''. Hence, we define more fine-grained categories, such as \texttt{InfoObjectLocAndOD} (information on object location and other details) in \teachda. These categories are designed in a way so that they could be re-purposed into broader dialog act category and intent/slot in the future by merging categories, if needed. As in a TOD, inform would be the DA tag, intent could be \texttt{inform\_object\_location} or \texttt{object\_location} could be slot category. 
Thus, we combine the use of many standardized dialog acts such as \texttt{Greetings},  \texttt{Acknowledge},  \texttt{Affirm / Deny} with domain-specific finer grained dialog acts replacing the typical \texttt{Inform} and \texttt{Request} dialog acts. 

Additionally, since the \teach dataset is not constrained by turn taking or a pre-defined dialog flow, sometimes a single utterance may perform multiple communicative functions.
To address this, similar to ~\citealt{Core1997CodingDW}, 
we allow multiple dialog acts per utterance and require annotators to mark utterance spans corresponding to each dialog act.

There exist other multimodal task-oriented dialog datasets that include annotations of dialog acts such as Situated and Interactive Multimodal Conversations (SIMMC 2.0) ~\cite{KotturMGD21} and Multimodal Dialogues (MMD) ~\cite{saha2018mmd}.  
These are multimodal datasets in the shopping domain that allows users to view products visually, and engage in dialog with an agent where the agent can take actions to refine the products available for the user to view. 
However, in contrast to the \teach dataset considered in our work, the dialogs are created by first simulating probable dialog flows and then having annotators paraphrase utterances. 
As such, in these datasets, utterances clearly map to predefined dialog acts and follow patterns expected by the designers. 
These may not fully cover the range of possible conversational flows that can happen between humans in an unconstrained multimodal context, as can be observed in \teach. 
The Human Robot Dialogue Learning (HuRDL) corpus includes annotations of human-human multimodal dialogs, with a focus on  classifying different types of clarification questions to be used by a dialog agent~\cite{gervits2021hurdl} but it is limited in size - consisting of only 22 dialogs, in contrast to the over 3,000 dialogs in \teach.  
Another related dataset is MindCraft~\cite{bara2021mindcraft} where annotators are periodically asked to answer questions in the middle of the collection of dialog sessions to elicit their belief states.  However, belief states do not map directly to utterances and do not directly capture communicative intents, differentiating them from dialog acts.

Prior works propose models for predicting dialog acts given the current utterance and
context~\cite{kalchbrenner2013recurrent,LeeD16,RibeiroRM19}, dialog acts of previous utterances 
or both~\cite{PaulGH19}. We perform similar experiments on our dataset to tag the dialog acts of given utterances and also to predict the dialog acts of future utterances. 
Due to the limited set of situated dialog datasets annotated with dialog acts, there has been relatively limited work on exploring the benefit of dialog acts on predicting an agent's future behavior in the environment. However, there are works that 
explore when to engage in a dialog as opposed to acting in the environment~\cite{GervitsTRS20,ChiSEKH20,visitron:2021}. 
While we do not directly model this problem, we experiment with the \teach Execution from Dialog History task, where the end of our predicted action sequence would signal the need for another dialog utterance.

\begin{table*}[t]
\tabcolsep 2pt
\centering
\small
\begin{tabular}{L{4cm}L{2.2cm}L{3.8cm}ccc}
\toprule
Dialog Act & Category & Example & Count & \commander (\%) & \follower (\%) \\
\midrule
\texttt{Instruction} & Instruction & fill the mug with coffee & 11019 & 99.4 & 0.6 \\
\texttt{ReqForInstruction} & Instruction & what should I do today?  & 4043 &0.7 & 99.3\\ 
\texttt{RequestOtherInfo} & Instruction & How many slices of tomato? &675& 0.75 & 99.25\\ 
\texttt{RequestMore} & Instruction & Is there anything else to do & 503 &0.2  &99.80\\

\texttt{InfoObjectLocAndOD} & Object/Location & knife is behind the sink & 6946& 99.4 & 0.6 \\
\texttt{ReqForObjLocAndOD} & Object/Location & where is the mug?  & 2010 & 0.3 & 99.70  \\ 
\texttt{InformationOther}& Object/Location & Mug is already clean  &1148 & 88.76 & 11.24 \\ 
\texttt{AlternateQuestions} & Object/Location & yellow or blue mug? & 123 &  27.65 & 72.35\\

\texttt{Acknowledge} &  Generic & perfect & 7421 & 21.38 & 78.62\\ 
\texttt{Greetings} &  Generic & hello & 2565 & 44.01 & 55.9 \\ 
\texttt{Confirm} & Generic & Should I clean the cup? & 726 & 25.75 & 74.25 \\ 

\texttt{MiscOther} & Generic &  ta-da & 607 & 52.22 & 47.78 \\ 
\texttt{Affirm} & Generic & Yes & 460 & 78.26 & 21.74 \\ 
\texttt{Deny} & Generic & No & 161 & 72.92 & 26.08\\ 

\texttt{FeedbackPositive} & Feedback & great job & 2745& 97.12& 2.88 \\ 
\texttt{FeedbackNegative} & Feedback & that is not correct &46 & 95.65 & 4.35\\

\texttt{OtherInterfaceComment} & Interface & Which button opens drawer & 486 & 60.09 & 39.91 \\

\texttt{NotifyFailure} & Interface & not able to do it & 408 & 3.68 & 96.32 \\

\bottomrule
\end{tabular}
\caption{Dialog act labels, total number of utterances and frequencies per speaker type in overall corpus.}
\label{tab:da_def_and_stats}
\end{table*}

\section{\teachda dataset}

The \teach dataset~\cite{padmakumar2021teach} consists of situated dialogs between human annotators 
role playing a user (\commander) and robot (\follower) collaborating to complete household tasks. 
In each dialog session, there is a high level task that the \follower\ is expected to accomplish, for example 
\task{Make Coffee} or \task{Prepare Breakfast}.  Details of the task are known to the \commander\ but 
not the \follower. The \follower\ needs to engage in a dialog with the user to identify the task to be 
completed, customize the task (for example identify what dishes need to be prepared for breakfast) 
or obtain additional information such as locations of relevant objects, or more detailed steps needed to accomplish a task, and translate these to actions that can be executed in a simulated environment to complete the task. 

In this work, we annotate the \teach dataset with dialog acts (we refer to this new, annotated dataset as \teachda) to better understand how language is used in task-oriented situated dialogs. We also explore the usefulness of these dialog acts 
to develop better agents that can converse in natural language and act in a situated environment for task completion.
The \teachda dataset consists of ~39.5k utterances from ~3,000 dialogs, 60\% of which are from the \commander\ and the rest from the \follower.

We find that other dialog act frameworks for multimodal datasets~\cite{gervits2021hurdl,KotturMGD21,saha2018mmd} tend to be domain specific and do not cover all utterance types that would be beneficial for embodied task completion. 
Hence, we propose a new set of dialog acts for embodied task completion based on the communicative functions we observe in the \teach dataset. 
Whenever possible, for utterances that are not very specific to the \teach task, we have borrowed dialog acts from prior work. These include dialog acts related to generic chit chat such as \texttt{Greetings}, \texttt{Affirm}, \texttt{Deny} and \texttt{Acknowledge}~\cite{PaulGH19}. 

In total, we defined 18 dialog acts that covered all utterances in \teach. Our careful analysis of utterances in \teach data lead to 5 broader categories of dialog acts as shown in Table ~\ref{tab:da_def_and_stats}.

\begin{itemize}[noitemsep]
    \item Generic: Acts that fall under conventional dialog such as opening and closing of dialog, 
    \item Instruction Related: Which represent the utterances related to actions that should be performed in the environment to accomplish the household task. 
    \item Object/Location related: Represents requests and information seeking utterances related to objects that need to be handled or manipulated for the specific \teach task. Many of these are on the specifics of object location (where to find it, where to place it) and queries on disambiguation related to objects or their locations. 
   \item Interface Related: Utterances related to \teach data annotation itself (\texttt{NotifyFailure} and \texttt{OtherInterfaceComment})
   \item Feedback related: Utterances used to provide feedback (both positive and negative) on navigation, object manipulation and in general task execution. 

\end{itemize}


\begin{figure*}[t]
    \centering
    \includegraphics[width=.99\textwidth]{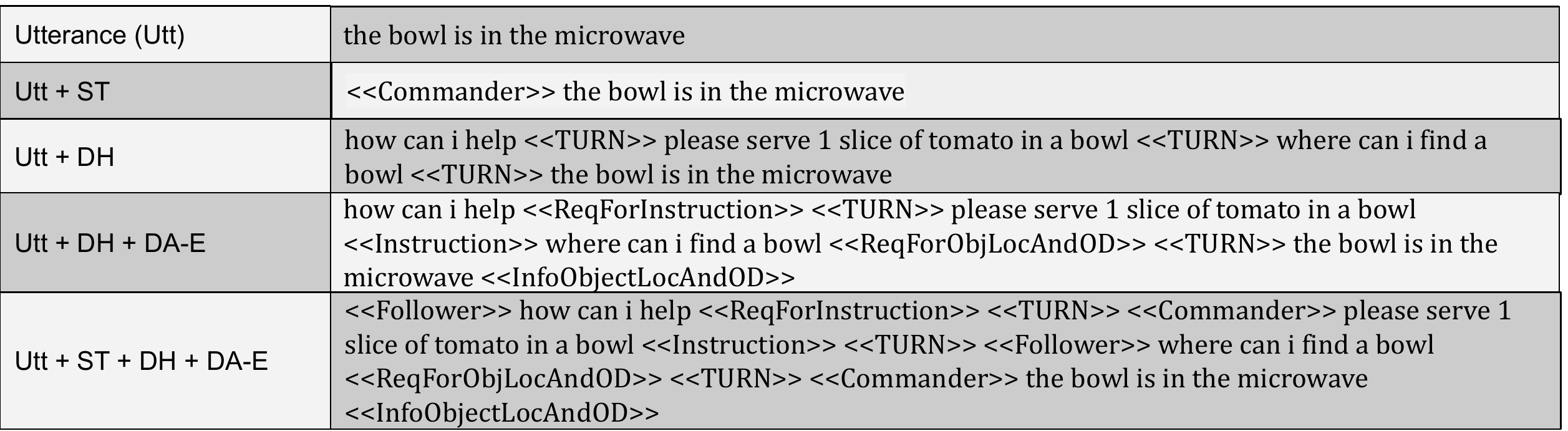}
    \caption{Sample input to dialog act prediction or next turn dialog act prediction models showing incorporation of speaker and dialog history} 
    \label{fig:da_pred_input}
\end{figure*}

We hired expert annotators who are fluent in English to annotate utterances from the \teach dataset with our dialog acts.
Annotators were shown the complete dialog and asked to annotate each utterance with the most appropriate 
dialog act. When an utterance had multiple dialog acts applicable, annotators were asked to 
divide the utterance into spans and annotate each span with a single dialog act label. 
We observed that 7\% of the utterances were segmented to have multiple dialog acts. To measure the quality of the annotations, 
on a small subset of ~235 utterances (17 dialogs), we collected annotations from two annotators. On this subset, we observed a Cohen's kappa score of 0.87.
We include an example \teach session in Figure~\ref{fig:example_figure_da} for the task \textit{Boil Potato} containing dialog act acctions for each utterance.

Similar to many task-oriented dialogs, we observe a strong correlation between the speaker role (\commander\ or \follower) and the dialog act of an utterance. For example, the majority of the inform utterances are from \commander\ i.e., where \commander\ gives instructions or informs object locations or other details on the task, whereas majority of the request utterances (instructions, object locations etc.)  are from \follower.
In Table~\ref{tab:da_def_and_stats}, we present the set of dialog acts, definitions and their frequency distributed across \commander\ and \follower\ utterances. 
We observe that some communicative functions such as clarification of ambiguity are relatively infrequent in this dataset. We group together such rare functions into a single dialog act \texttt{MiscOther}.

\begin{table}[t]
    \centering
    \tabcolsep 1.3pt
    \begin{tabular}{p{3.5cm}cccc}
    \toprule
       & Valid & Valid & Test & Test \\
       & seen & unseen & seen & unseen \\
       \midrule
        Utterance   & 85.59 & 83.74 & 85.88 & 83.59 \\
        +Speaker Tags (ST) & 87.98 & 85.91 & 87.55 & 85.73 \\
        + Dialog History (DH)  & 86.7 & 84.66 & 86.48 & 84.25\\
        +  DH + DA-E & \bf{88.6} & \bf{86.32} & 88.35 & \bf{86.09}\\
        + DH + ST + DA-E & 88.35 & 86.15 & \bf{88.54} & 85.89\\
        
\midrule 
\multicolumn{5}{c}{\follower\ utterances only}  \\
\midrule 
        Utterance  & 83.12 & 79.58 &  84.86 & 83.85 \\
        +Speaker Tags (ST) & 86.84 &  82.26 &  88.33 & \bf{87.71}\\
        +Dialog History  (DH) & 86.52 &84.13 &  86.67 & 84.53\\
        + DH +DA-E & \bf{88.62} & \bf{85.87} & 88.82 & 86.56\\
        + DH + ST + DA-E & 88.32 & 85.79 & \bf{89.22} & 86.3\\

\midrule 
\multicolumn{5}{c}{\commander\ utterances only}  \\
\midrule
        Utterance   &  87.16 & 86.71 & 86.5 & 83.42 \\
        + Speaker Tags (ST) &  \bf{88.70} & \bf{88.52} & \bf{87.08} & 84.42\\
        + Dialog History (DH) & 87.11 & 81.03 & 85.79 & 83.49\\
        + DH + DA-E & 88.55 & 87.90 & 86.69 & \bf{84.84} \\
        + DH + ST + DA-E & 88.42 & 87.4 & 86.15 & 84.79\\
        \bottomrule
        \end{tabular}
    \caption{Dialog Act prediction accuracy scores for whole \teachda dataset. We also report accuracy scores for \follower\ and \commander\ utterances separately. }
    \label{tab:da_classification}
\end{table}


\section{Experiments}

In this section, we explore how dialog acts can be used for various modeling tasks including predicting the agent's future behavior in the environment.
We explore the following tasks (i) dialog act classification: predicting the dialog act of an utterance; (ii) future turn dialog act prediction given dialog history; 
(iii) given \teach dialog history, predicting a plan for the task and (iv) given dialog history and the past actions in 
environment, predicting the entire sequence of low-level actions to be executed in the \teach environment to complete the task (Execution from Dialog History (EDH) benchmark from \citealt{padmakumar2021teach}). 
Note that \teach has two validation and two test splits each - seen and unseen. These refer to visual differences between the environments in which gameplay sessions occurred. With the exception of the EDH experiment, since we only focus on language, we do not expect significant differences between the seen and unseen splits.

\subsection{Dialog Act Classification}
\label{ssec:expt_da_classification}

Dialog Act classification is the task of identifying the general intent of the user utterance in a dialog. 
While dialog act classification has been well explored in both task-oriented dialogs and 
open-domain dialogs, it is still an under explored problem in human-robot 
dialogs \cite{GervitsTRS20}. We study the \teach dataset to predict the dialog act for a 
given utterance. We experimented with fine-tuning a large pre-trained language 
model \textit{RoBERTa-base} for the classification of dialog acts\footnote{We also experimented with \textit{BERT-base} and \textit{TOD-BERT} but observed \textit{RoBERTa-base} 
performed consistently better}. 
We expect the speaker role (\follower\ or \commander) and the dialog context to be important for predicting the intent of an utterance.
To test this, we predict dialog acts with different input formats (shown in Figure~\ref{fig:da_pred_input}) ablating the value of speaker and context information (DH: all the previous utterances in the dialog, ST: speaker tags, DA-E: ground-truth dialog act tags of all the previous utterances in the dialog). We present our results in Table~\ref{tab:da_classification}. Similar to prior studies on dialog 
act classification for task-oriented dialogs, we observe that both the speaker tags and 
dialog history help in predicting the correct dialog act for a given utterance, and the best performance is observed when both of them are used. 

In \teach, the distribution of dialog acts varies with the speaker role (\commander\ vs. \follower) as shown in Table~\ref{tab:da_def_and_stats}. 
To understand the accuracy of the models on utterances of each speaker role, we also present results separated by speaker role in Table~\ref{tab:da_classification}.
We observed that both speaker tags and dialog history with previous turn dialog acts 
helped identifying dialog acts for \follower\ utterances. For \commander\ utterances both speaker tags and dialog history gave marginal improvements.


\begin{table}[t]
\tabcolsep 1.8pt
    \centering
    \begin{tabular}{p{3.5cm}cccc}
        \toprule
        & Valid & Valid & Test & Test \\
       & seen & unseen & seen & unseen \\
       \midrule
        DH  & 42.62 & 42.44 &  43.55 & 41.07\\
        DH + ST & 56.23  & 54.68  &  54.69 & 53.27\\ 
        DH + DA-E & 56.05 & 55.58 & \bf{56.49} & 53.45 \\
         DH + ST + DA-E & \bf{56.72} & \bf{56.14} & 56.28 & \bf{54.99}\\
        \midrule 
        \multicolumn{5}{c}{\follower\ utterances only}  \\
        \midrule 
        DH & 30.73 & 28.64 &  31.41 & 29.06\\
        DH +ST & 51.67  &  49.3 &  54.11 &   52.34\\
        DH + DA-E & 50.19 & 50.28 & \bf{54.72} & 52.24\\
         DH + ST + DA-E & \bf{52.17} & \bf{50.35} & \bf{54.72} & \bf{53.44} \\
        \midrule
        \multicolumn{5}{c}{\commander\ utterances only} \\
        \midrule
        DH & 49.27 & 51.08 & 50.07  & 48.08\\
        DH + ST & 58.78  & 58.05  & 55.01 &  53.82\\ 
        DH + DA-E & 59.33 & 58.9 & \bf{57.4} & 54.16\\
         DH + ST + DA-E & \bf{59.26} & \bf{59.77} & 57.11 & \bf{55.89}\\
        \bottomrule
    \end{tabular}
    \caption{Predict next utterance Dialog Act given dialog history. We also report results when next utterance is \commander\ and \follower\ separately. Speaker Tags: Additional to current utterance speaker tag we also provide next utterance speaker information.}
    \label{tab:next_turn_da_prediction}
\end{table}

\subsection{Next Dialog Act Prediction}
\label{ssec:expt_next_da_pred}

In end-to-end dialog models, predicting the desired dialog act for the next turn is useful for response generation~\cite{TanakaTA19}. 
Predicting the dialog act of the next response in \teach will provide insights into a model's ability to provide appropriate dialog responses. 
This is particularly useful for \follower\ utterances to enable the agent to identify when to ask for more instructions or additional information to accomplish a sub-task. 
We modeled this as a classification task where we provide dialog history until a particular turn as input and predict the dialog act of the next turn. 
In addition to providing dialog history, we also tested this to see if providing next turn speaker information will improve the performance of the model. 
Similar to our dialog act classification model in Section ~\ref{ssec:expt_da_classification} we fine-tuned a \textit{RoBERTa-base} model for predicting the dialog act of the next utterance. In Table ~\ref{tab:next_turn_da_prediction}, we present results for next dialog act prediction. We observe a significant improvement in the performance for next dialog act prediction when the next utterance is from the \follower\ and the speaker information or previous utterances dialog act is added to the input.
We hypothesize that the accuracy in this task is low compared to similar tasks in other task-oriented dialog datasets because this dataset does not enforce turn taking. The \commander\ or \follower\ may break up a single intent into multiple utterances and one may anticipate the next response from the other before it is asked. For example, if the \commander\ has asked the \follower\ to slice a tomato, the \commander\ may expect that the \follower\ is likely to then ask for the locations of the tomato or the knife and may start providing this information before the \follower\ has asked for it.  
Further, the \commander\ or \follower\ may have responded directly to visual cues or actions taken by the other in the environment. Hence, visual or environment information is likely also important for predicting future dialog acts. 

\begin{figure*}[t]
    \centering
    \includegraphics[width=.99\textwidth]{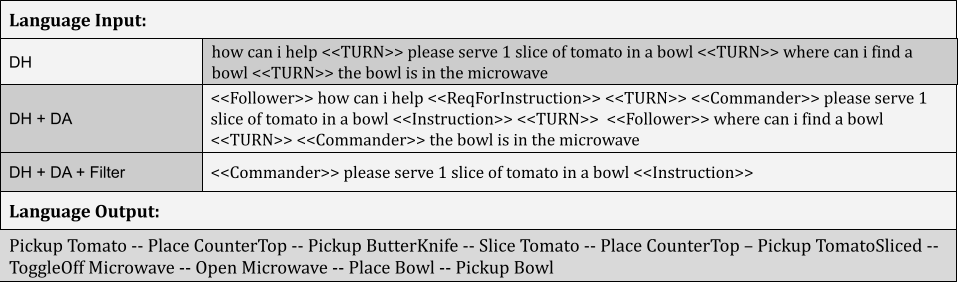}
    \caption{Sample input and output for plan prediction showing incorporation of speaker and dialog act information.} 
    \label{fig:plan_pred_input}
\end{figure*}

\begin{table*}[]
    \centering
    \tabcolsep 2pt
    \begin{tabular}{L{1.25cm} cccc c cccc c cccc}
    \toprule
    \multicolumn{15}{c}{\texttt{Game-to-Plan}} \\
    \midrule
    & \multicolumn{4}{c}{Percentage of valid plans} && \multicolumn{4}{c}{Plan tuple precision} && \multicolumn{4}{c}{Plan tuple recall} \\
    \cline{2-5} \cline{7-10} \cline{12-15} \\
    & Valid  & Valid  & Test  & Test  && Valid  & Valid  & Test & Test  && Valid  & Valid  & Test  & Test  \\
    & seen & unseen & seen & unseen && seen & unseen & seen & unseen && seen & unseen & seen & unseen \\
    \cline{2-5} \cline{7-10} \cline{12-15} \\
DH &  24.31 & \textbf{30.39} & \textbf{28.18} & 28.69 && 72.67 & \textbf{73.93} & 73.48 & \textbf{78.53} && 37.06 & \textbf{34.35} & 37.46 & \textbf{36.00} \\
+ DA &  25.97 & 23.86 & 19.89 & 26.83 && \textbf{75.29} & 73.0 & \textbf{74.81} & 77.52 && \textbf{38.18} & 33.7 & \textbf{39.28} & 35.31 \\
+ Filter  & \textbf{37.57} & 29.41 & 27.62 & \textbf{32.94} && 71.29 & 70.94 & 69.80 & 75.45 && 34.33 & 31.61 & 35.45 & 33.42 \\
    \midrule
    \multicolumn{15}{c}{\texttt{Dialog-History-to-Plan}} \\
    \midrule
DH & 23.76    & 23.69    & 25.41    & 24.45   & & 72.97    & \bf{73.47}    & \textbf{75.65}    & \textbf{78.64}   & & 36.38    & 34.06    & \textbf{39.11}    & \textbf{36.53} \\
+ DA & 24.31    & \textbf{30.39}    & \textbf{28.18}    & \textbf{28.69}   & & 72.67    & 73.93    & 73.48    & 78.53   & & \textbf{37.06}    & \textbf{34.35}    & 37.46    & 36.0  \\
+ Filter & \bf{26.52}    & 23.69    & 25.41    & 28.01   & & \textbf{73.66}    & 69.88    & 71.67    & 74.33   & & 36.08    & 31.29    & 35.83    & 33.12 \\
    \bottomrule
    \end{tabular}
    \caption{Plan prediction results. Using dialog act information helps increase the fraction of valid generated plans but not as much with plan precision or recall.}
    \label{tab:plan_prediction}
\end{table*}

\begin{figure*}[t]
    \centering
    \includegraphics[width=.99\textwidth]{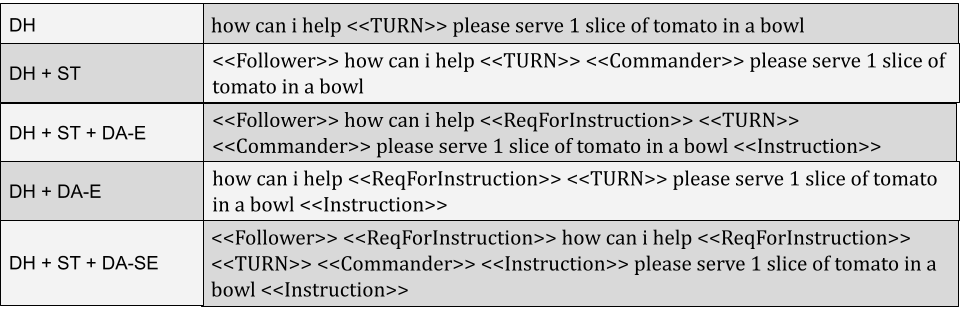}
    \caption{Language Input Variants for EDH.} 
    \label{fig:edh_input}
\end{figure*}

\begin{table*}[t]
    \centering
    \tabcolsep 1.8pt
    \begin{tabular}{L{2.4cm} rr r rr r rr r rr}
        \toprule
         & \multicolumn{5}{c}{\textbf{EDH Validation}} & \multicolumn{5}{c}{\textbf{EDH Test}} \\
         & \multicolumn{2}{c}{\textit{Seen}} & & \multicolumn{2}{c}{\textit{Unseen}} & & \multicolumn{2}{c}{\textit{Seen}} & & \multicolumn{2}{c}{\textit{Unseen}} \\
        \midrule
         Language Input & \multicolumn{1}{c}{SR \small{[TLW]}} & \multicolumn{1}{c}{GC \small{[TLW]}} && \multicolumn{1}{c}{SR \small{[TLW]}} & \multicolumn{1}{c}{GC \small{[TLW]}} && \multicolumn{1}{c}{SR \small{[TLW]}} & \multicolumn{1}{c}{GC \small{[TLW]}} && \multicolumn{1}{c}{SR \small{[TLW]}} & \multicolumn{1}{c}{GC \small{[TLW]}} \\
         \midrule 
         DH & 7.9 \small{[1.0]}	& 7.1 \small{[3.3]}	&& 6.7 \small{[0.4]}	& 3.9 \small{[1.5]}	&& 10.5 \small{[0.5]}	& 7.9 \small{[3.2]} && 7.5 [0.7] & 5.6 [1.9] \\
         + ST & 6.7 \small{[0.5]} & 7.4 \small{[2.8]} && 6.7 \small{[0.8]} & 4.0 \small{[1.5]} && 9.8 \small{[0.9]} & 8.3 \small{[2.9]} && 7.1 [0.8] & 6.6 [1.7] \\
         
        + DA-E & 8.5 [0.6] & \textbf{8.2 [3.3]} && 6.7 [0.5] & \textbf{5.0 [1.9]} && \textbf{12.2 [1.2]} & 8.6 [3.7] && 7.4 [0.8] & 6.1 [2.3] \\
        + DA-SE & 7.8 [1.8] & 6.4 [4.0] && 7.2 [0.6] & 4.6 [1.6] && 11.0 [0.7] & \textbf{10.1 [4.3]} && \textbf{7.7 [0.8]} & 6.2 [1.8] \\
        + ST + DA-SE & \textbf{8.7 [1.0]} & 7.3 [2.6] && \textbf{7.5 [0.8]} & 4.4 [1.8] && 9.9 [0.7] & 8.0 [2.9] && 7.0 [0.7] & \textbf{7.2 [2.2]} \\
         \bottomrule
    \end{tabular}
    \caption{We experiment whether addition of speaker or dialog act information improves performance of the Episodic Transformer (E.T.) model on the Execcution from Dialog History (EDH) task. In most cases, speaker information is not found to be beneficial but adding dialog acts at the end or start and end of an utterance is seen to provide small improvements in performance.}
    \label{tab:expt_edh}
\end{table*}

\subsection{Plan Prediction}
\label{ssec:expt_plan_prediction}

In robotics, task planning is the process of generating a sequence of symbolic actions to 
guide high-level behavior of a robot to complete a task~\cite{ghallab:16}. 
In this experiment, we consider a simple plan representation where a task plan consists of 
a sequence of object manipulations that need to be completed in order for the task to be successful. 
An example is included in Figure \ref{fig:plan_pred_input} 
When executing such a plan, the robot will need to navigate to required objects and additional 
steps may be required based on the state of the environment (for example if the microwave is too full, 
the robot may need to partially clear it first).

However, it should be possible to generate the plan for a task based on the dialog alone. 
We explore two settings for this
\begin{itemize}[noitemsep]
    \item \texttt{Game-to-Plan}: Given the entire dialog from a gameplay session, 
    predict the plan - that is, all object interaction actions taken during that gameplay session.
    \item \texttt{Dialog-History-to-Plan}: Given a portion of dialog history from a gameplay session, 
    predict the object interaction actions that need to occur until the next dialog utterance.
\end{itemize}
The \texttt{Game-to-Plan} setting is more likely to be useful for post-hoc analysis of such 
situated interactions after they have occurred, whereas the \texttt{Dialog-History-to-Plan} 
setting can be used to build an embodied agent that engages in dialog with a user and 
executes actions in a virtual environment based on information obtained in the dialog. 
At any point in time, such an agent would predict the next few object interactions 
to be accomplished given the dialog history so far, complete them and then use 
another module that makes use of subsequent dialog act prediction 
(section \ref{ssec:expt_next_da_pred}) to engage in further dialog with the user.

We model plan prediction as a sequence to sequence task where the input consists 
of the dialog / dialog history, and the output as a sequence of alternating 
object interaction actions (eg: \texttt{Pickup}, \texttt{Place}, \texttt{ToggleOn}) 
and object types (eg: \texttt{Mug}, \texttt{Sink}).  
We experiment with augmenting the dialog history with dialog act information 
(+ DA information) and filtering the input dialog to only contain utterance 
segments annotated as being of type \texttt{Instruction} (+ filter) 
We fine-tune a \texttt{BART-base} model for this task and evaluate different 
experimental conditions on the following metrics:
\begin{itemize}[noitemsep]
    \item Fraction of valid plans: Fraction of generated output 
    sequences that consist of alternating valid actions and object types.
    (For example \texttt{(Pickup, Mug), (Place, Sink) (ToggleOn, Faucet)} is a valid 
    sequence while \texttt{(Pickup, Mug) (Sink) (ToggleOn, Faucet)} and 
    \texttt{(Pickup, Mug) (Place) (ToggleOn, Faucet)}) are not due to the missing action for \texttt{Sink} and the missing object for \texttt{Place} respectively. 
    \item Precision of (action, object) tuples: We identify a valid object type 
    followed by a valid action as an (action, object) tuple and precision is the 
    fraction of such tuples in the generated output present in the ground truth plan.
    \item Recall of (action, object) tuples: Recall is the fraction of (action, object) 
    tuples in the ground truth plan present in the generated output. 
\end{itemize}

The results are included in Table~\ref{tab:plan_prediction}. 
We notice that addition of dialog act information and filtering to relevant dialog acts improves performance in some splits but not others. More improvements are seen in the \texttt{Dialog-History-to-Plan} setting compared to the \texttt{Game-to-Plan} setting.
We hypothesize that this is because the model is able to automatically identify the dialog act from the utterance text and hence does not need it to be explicitly specified.

\subsection{Execution from Dialog History}
\label{ssec:expt_edh}

The Execution from Dialog History (EDH) task defined in the \citealt{padmakumar2021teach} 
is an extension of the above task. Instead of simply predicting important object interactions, 
given dialog history and past actions in the environment, a model is expected to predict a 
full sequence of low level actions to accomplish 
the task described in the dialog. 
Action sequences predicted by the model are executed in the virtual environment and models are evaluated based on how many required object state changes are accomplished.
The metrics used for this task include the fraction of successful state changes (goal condition 
success rate or GC), the fraction of sessions for which all state changes 
were accomplished (success rate or SR) and Trajectory Length Weighted versions 
of these metrics that multiply the metrics with the ratio of the ground truth path 
length to the predicted path length - where a lower value of the trajectory weighted 
metric suggests that the model used longer sequences of actions to accomplish the same state changes. 

We borrow the Episodic Transformer (E.T.) model proposed in \citealt{padmakumar2021teach} and vary the language 
input (with a baseline of just the dialog history (DH)) by adding speaker tags (+ST) and ground-truth dialog act tags at the start (+DA-S), end (+DA-E) or both (+DA-SE).
We present the results for selected set of experiments in Table~\ref{tab:expt_edh}. We observe small performance improvements on success rate of up to 2 points when the language input is marked up with dialog acts, either at the end or start and end of an utterance, 
but less benefit is observed from speaker information. We believe that stronger 
improvements will likely be observed when using a more modular approach (eg: \cite{min2021film}) 
where it is easier to decouple the effects of errors arising from language understanding 
from those arising from navigation which is the most difficult component when predicting such low-level actions~\cite{blukis2022persistent,jia2022learning,min2021film}. 

\section{Conclusion}
\label{sec:conclusion_future_work}

We propose a new dialog act annotation framework for embodied task completion dialogs 
and use this to annotate the \teach dataset - a dataset of over 3,000 unconstrained, situated 
human-human dialogs. We evaluate baseline models for predicting dialog acts of utterances, 
demonstrate that predicting future dialog acts from past ones is much more difficult in 
dialog datasets that are not constrained by turn taking.
Towards guiding agent actions in the environment beyond dialog, we show explore the benefit of dialog acts in the generation of plans, and improve end-to-end performance in the \teach Execution from Dialog History task.

\section{Future Work}

Unlike the majority of dialog datasets, situated or otherwise, utterances in the \teach dataset are not constrained by a pre-designed dialog act schema or by turn taking. We observe that this makes it much more difficult than expected to predict subsequent dialog acts given past ones - the predictability of which has been typically used to design 
dialog simulators~\cite{schatzmann2009hidden,keizer2010parameter}. We believe that annotation of this large and more natural dataset will aid in the development of
more realistic dialog simulators, which can in turn result in the development of 
more natural dialog agents. 
Further, in \teach, visual cues or actions taken by the agent in the environment might play an important role for predicting future dialog acts. This would be an interesting direction to explore for future. 
Finally, we hypothesize that there is considerable scope in using such annotated dialog acts to develop modular models for embodied task completion that involve better language 
understanding, and to generate realistic situated dialogs for data augmentation.

\bibliography{main}
\bibliographystyle{acl_natbib}

\appendix

\begin{table*}[t]
\tabcolsep 2pt
\centering
\small
\begin{tabular}{L{4cm}L{4cm}L{7.5cm}}
\toprule
Dialog Act & Task & Agent: Example  \\
\midrule
\multirow{3}{*}{\texttt{Instruction}} & Water Plant &  \commander: The plant by the sink needs to be watered \\
& Plate Of Toast & \commander: please slice bread and toast 1 slice\\
& Plate Of Toast & \commander: lets make a slice of toast\\
\midrule

\multirow{3}{*}{\texttt{InfoObjectLocAndOD}} &  Plate Of Toast & \commander: knife is in the fridge\\
& Plate Of Toast & \commander: the clean plate is on the white table \\
& Clean All X & \commander: right cabinet under the sink\\
\midrule

\multirow{3}{*}{\texttt{Acknowledge}} &  Make Coffee & \commander: we are done! \\
& Clean All X & \follower: Plate is clean\\
& N Slices Of X In Y &\follower: found it\\
\midrule

\multirow{3}{*}{\texttt{ReqForInstruction}} & Put All X On Y & \follower: how can I help\\
& Put All X On Y & \follower: what are my directions\\
& Plate Of Toast & \follower: what is my task today\\
\midrule

\multirow{ 3}{*}{\texttt{FeedbackPositive}} & Plate Of Toast & \commander: good job\\
& Put All X In One Y &  \commander: that's it good job \\
& Water Plant & \commander: thank you its seems to be done\\
\midrule

\multirow{ 3}{*}{\texttt{Greetings}} &  Make Coffee & \commander: Hi how are you today?\\
& Water Plant & \follower: Good day \\
& Boil X & \commander: Good morning\\
\midrule

\multirow{ 3}{*}{\texttt{ReqForObjLocAndOD}} & Clean All X & \follower: where is the dirty cookware?\\
& Plate Of Toast & \follower: Can you help me find knife? \\
& Put All X In One Y & \follower: where is the third one? \\
\midrule

\multirow{3}{*}{\texttt{InformationOther}}& Make Coffee & \commander: Don't take martini glass\\
& Boil X& \commander: You keep walking past them \\
& Boil X & \commander: That looks cooked already \\
\midrule

\multirow{3}{*}{\texttt{Confirm}} & Put All X In One Y & \follower: was that everything\\
& Salad & \commander: you can see the toaster right? \\
& N Slices of X in Y& \follower: Shall I turn off the water?\\
\midrule

\multirow{3}{*}{\texttt{RequestOtherInfo}} & Breakfast & \follower: how many slices of each?\\
& Clean All X & \follower: what pieces? \\
& Plate Of Toast & \follower: shall i take it to the toaster now\\
\midrule

\multirow{3}{*}{\texttt{MiscOther}} &Sandwich &  \commander: One sec\\
& Salad & \commander: Common!! \\
& Breakfast & \commander: Thant's my bad...Sorry \\
\midrule

\multirow{3}{*}{\texttt{RequestMore}} & N Cooked Slices Of X In Y & \follower: Is there anything more I can help with? \\
& Salad & \follower: what else would you like me to do \\
& Clean All X & \follower: Any more tasks? \\
\midrule

\multirow{3}{*}{\texttt{OtherInterfaceComment}} & Plate of Toast & \follower: Finish and report a bug? \\
& Clean All X & \follower: refresh the page \\
& Put All X On Y  & \follower: connection is slow \\
\midrule

\multirow{3}{*}{\texttt{Affirm}} & Water Plant & \commander: yes, you can use the green cup \\
& Breakfast &  \commander: yes, toast the bread\\
& Put All X On Y & \commander: yes please\\
\midrule

\multirow{ 3}{*}{\texttt{NotifyFailure}} & Make Coffee & \follower: It's not turning on the coffee. \\
& N Slices Of X In Y & \follower: tomato won't fit in those \\
& Sandwich & \follower: can't seem to grab the knife in cabinet \\
\midrule

\multirow{ 3}{*}{\texttt{Deny}} & Make Breakfast & \commander: No don't toast the bread\\
& Salad & \commander: don't\\
& Plate of Toast & \commander: don't think so\\
\midrule

\multirow{ 3}{*}{\texttt{AlternateQuestions}} & N Cooked Slices Of X In Y & \follower: Do I boil it or slice it? \\
& Clean All X & \follower: To the left or right of the stove? \\
& Make Coffee & \follower: This mug or the other one? \\
\midrule

\multirow{ 3}{*}{\texttt{FeedbackNegative}} & Make Coffee & \commander: you don't have the correct mug \\
& N cooked Slices of X in Y & \commander: task not complete \\
& Plate of Toast & \commander: wrong plate \\

\bottomrule
\end{tabular}
\caption{Example utterances for Dialog act labels that could be observed in different \teach tasks from \commander and \follower.}
\label{tab:da_examples_task}
\end{table*}

\section{Further Experiment Details}

\subsection{Dialog Act Classification and Next Turn Dialog Act Prediction}

Both for dialog act classification and next turn dialog act prediction models, 
we finetune a \texttt{RoBERTa-base} model 
for multiclass classification with 18 classes (our target number of dialog acts). 
For all the experiments were run using Huggingface library and the publicly available pre-trained models.
Additional to the utterance we provide dialog-context and speaker information (mentioned as dialog history (DH) and Speaker Info (SI)) and train the classifiers for a maximum sequence length of 512 tokens. When the input exceeds 512 tokens we truncate from left i.e., we keep the most recent context. We use a batch size of 16 per GPU and accumulate gradients across 4 GPU instances. We use a learning rate of $2e-05$ and train for 5 epochs.

\subsection{Plan Prediction}

For the plan prediction task, we finetune a \texttt{bart-base} model, treating the problem as sequence to sequence prediction. A sample input and output from the \texttt{Game-to-Plan} version of the task are included below:

Sample Input:
\begin{verbatim}
what do I do? <<TURN>> making 
coffee <<TURN>> grab a mug 
<<TURN>> where is tyhe mug? 
<<TURN>> on the counter next to 
you <<TURN>> empty, and wash 
<<TURN>> should I wash the mug 
<<TURN>> place in coffee maker 
after cleaning <<TURN>> yes 
<<TURN>> okay <<TURN>> turn on 
water <<TURN>> turn off <<TURN>> 
place in coffee maker next to 
sink <<TURN>> empty first 
<<TURN>> turn on <<TURN>> great 
job....we're done... <<TURN>> 
\end{verbatim}

Sample Output:
\begin{verbatim}
Pickup Mug Pour SinkBasin Place 
SinkBasin ToggleOn Faucet 
ToggleOff Faucet Pickup Mug Pour 
SinkBasin Place CoffeeMachine 
ToggleOn CoffeeMachine
\end{verbatim}

Note that we do not include any punctuation in the output sequence to demarcate (action, object) tuples and instead post process the generated sequence deleting any action not followed by an object or object not preceded by an action for evaluation. Also, while we use \texttt{$\langle\langle$TURN$\rangle\rangle$} in the above example to demarcate turns, in actual implementation, the default BART separator token is used. 

All experiments are run using the HuggingFace library and pretrained models~\footnote{\url{https://huggingface.co/}}. We use a batch size of 2 per GPU accumulating gradients from batches on 4 GPUs of an AWS `p3.8xlarge` instance leading to an effective batch size of 8. Training was done for 20 epochs. We use the AdamW optimizer with $\beta_1=0.9$, $\beta_2=0.99, \epsilon=1e-08$ and weight decay of 0.01. We use a learning rate of $5e-05$ with a linear warmup over 500 steps. Where necessary, we right-truncate the input to the model's limit of 1024 tokens as we believe that when an incomplete conversation must be used, the model may be able to infer most of the necessary steps from the task information which is likely to be indicated by the first few utterances of the conversation. 

The primary hyperparameter tuning we experimented with involved the position at which the dialog act was inserted relative to the utterance, which was one of
\begin{itemize}
    \item START$\_$OF$\_$SEGMENT - Start of the utterance segment 
    \item END$\_$OF$\_$SEGMENT - End of the utterance segment 
    \item START$\_$END$\_$SEGMENT - Start and end of the utterance segment 
\end{itemize}
and the format used to insert dialog act information, which was one of 
\begin{itemize}
    \item NO$\_$CHANGE$\_$TEXT - The name of the dialog act is inserted in Camel case as a part of the input text to the model.
    \item FILTER - Retain only utterances marked with the dialog act \task{Instruction}. Additionally, the name of the dialog act is inserted in Camel case as a part of the input text to the model.
    \item TAGS$\_$IN$\_$TEXT - The name of the dialog act in Camel case is surrounded by $\langle\langle\rangle\rangle$.
    \item TAGS$\_$SPL$\_$TOKENS - The name of the dialog act in Camel case is surrounded by $\langle\langle\rangle\rangle$ and this is specified as being a special token so that it does not get split by the tokenizer.
    \item SPLIT$\_$WORDS$\_$TEXT - The name of the dialog act is split into individual words (for example, \task{RequestForInstruction} becomes ``request for instruction'') and these are inserted into the text.
\end{itemize}
We also tuned whether speaker information was passed to the model. None of the format, position or speaker tag choices were found to consistently outperform the other.

For the DH rows in table \ref{tab:plan_prediction}, neither the position, nor the format of dialog acts is relevant as no dialog act information is used. We also do not filter utterances. The best +DA row in the \texttt{Game-to-Plan} setting used dialog acts in format SPLIT$\_$WORDS$\_$TEXT in position END$\_$OF$\_$SEGMENT with speaker tags. The best +Filter row in the \texttt{Game-to-Plan} setting used dialog acts in format START$\_$END$\_$SEGMENT without speaker tags. The best +DA row in the \texttt{Dialog-History-to-Plan} setting used dialog acts in format SPLIT$\_$WORDS$\_$TEXT in position START$\_$OF$\_$SEGMENT without speaker tags. The best +Filter row in the \texttt{Dialog-History-to-Plan} setting used dialog acts in format END$\_$OF$\_$SEGMENT without speaker tags.

\subsection{Execution from Dialog History}

We adapt the Episodic Transformer (E.T.) model first introduced in \cite{pashevich2021et} and used for baseline experiments in \cite{padmakumar2021teach} on the TEACh dataset. We keep all training parameters constant from \cite{padmakumar2021teach} and primarily experiment with the input format as described in the main paper. Unlike our previous experiments, since the language encoder of the E.T. model is trained from scratch using only the vocabulary present in the training data, we insert dialog acts and speaker indicators as individual tokens in the input that will be treated identically to other text tokens.

\section{Dialog Acts}

In Table~\ref{tab:da_examples_task} we add further examples for each dialog act (for both \follower and \commander) from different \teach tasks to demonstrate the difference in type of utterances we observe in the dataset.

\end{document}